\documentclass[preprint,journal]{vgtc}       

\ieeedoi{10.1109/TVCG.2019.2934619}
\usepackage{mathptmx}
\usepackage{graphicx}
\usepackage{times}
\usepackage{subcaption}

\usepackage[bookmarks,backref=true,linkcolor=black]{hyperref} 
\hypersetup{
  pdfauthor = {},
  pdftitle = {},
  pdfsubject = {},
  pdfkeywords = {},
  colorlinks=true,
  linkcolor= black,
  citecolor= black,
  pageanchor=true,
  urlcolor = black,
  plainpages = false,
  linktocpage
}

\onlineid{1071}

\vgtccategory{System}

\title{The What-If Tool: Interactive Probing of Machine Learning Models}

\author{James Wexler, Mahima Pushkarna, Tolga Bolukbasi, Martin Wattenberg, Fernanda Vi\'egas, and Jimbo Wilson}
\authorfooter{
\item
 James Wexler, Mahima Pushkarna, Tolga Bolukbasi, Martin Wattenberg, Fernanda Vi\'egas, and Jimbo Wilson are with Google Research. E-mail: {jwexler,mahimap,tolgb,wattenberg,viegas,jimbo}@google.com.
}

\shortauthortitle{Wexler \MakeLowercase{\textit{et al.}}: The What-If Tool: Interactive Probing of Machine Learning Models}

\abstract{A key challenge in developing and deploying Machine Learning (ML) systems is understanding their performance across a wide range of inputs. To address this challenge, we created the \textit{What-If Tool}, an open-source application that allows practitioners to probe, visualize, and analyze ML systems, with minimal coding. The What-If Tool lets practitioners test performance in hypothetical situations, analyze the importance of different data features, and visualize model behavior across multiple models and subsets of input data. It also lets practitioners measure systems according to multiple ML fairness metrics. We describe the design of the tool, and report on real-life usage at different organizations.
} 


\CCScatlist{ 
 \CCScat{K.6.1}{Management of Computing and Information Systems}%
{Project and People Management}{Life Cycle};
 \CCScat{K.7.m}{The Computing Profession}{Miscellaneous}{Ethics}
}

  \teaser{
\begin{subfigure}{.33\textwidth}
  \captionsetup{width=0.8\textwidth}
  \vspace*{5mm}
  \centering 
  \includegraphics[width=0.9\linewidth]{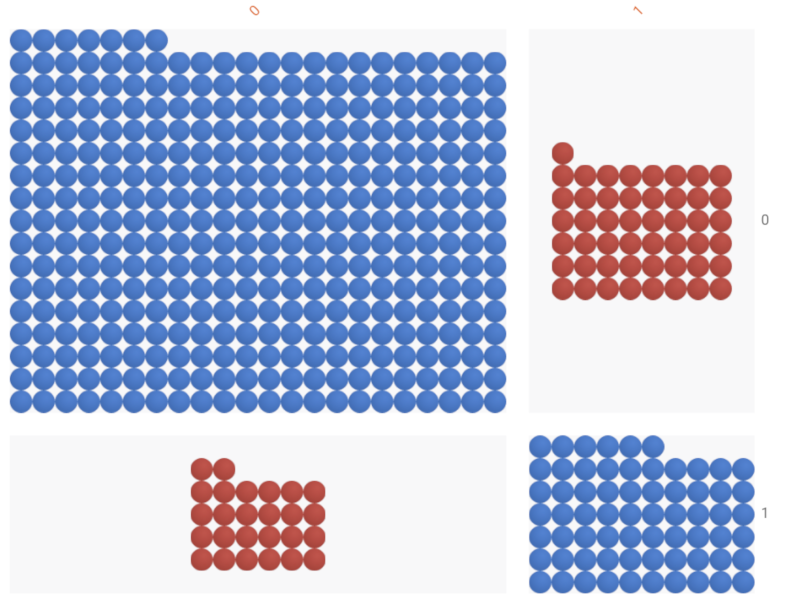}
  \caption{Confusion matrix of a single binary classification model, colored by prediction correctness}
  \label{fig:confMatrix}
\end{subfigure}
\begin{subfigure}{.33\textwidth}
  \captionsetup{width=0.8\textwidth}
  \centering 
  \includegraphics[width=0.9\linewidth]{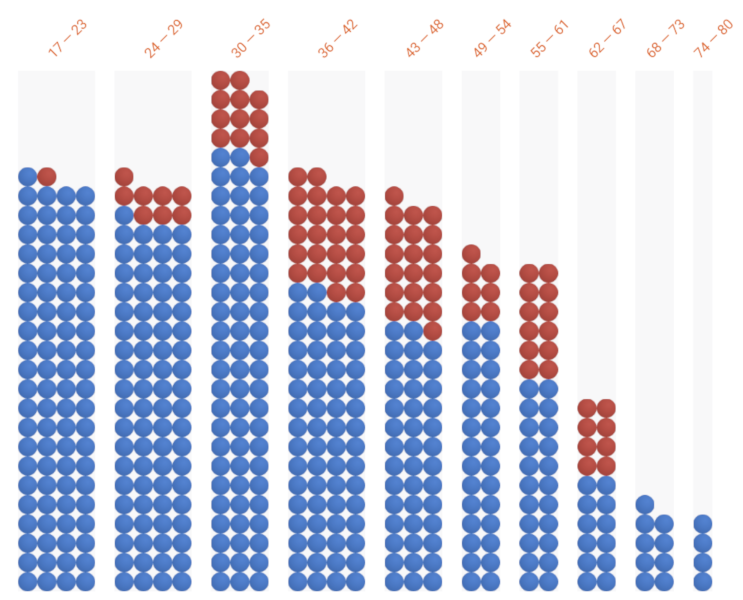}
  \caption{Histogram of age, colored by classification}
  \label{fig:ageHistogram}
\end{subfigure}
\begin{subfigure}{.33\textwidth}
  \captionsetup{width=0.8\textwidth}
  \vspace*{3mm}
  \centering 
  \includegraphics[width=0.9\linewidth]{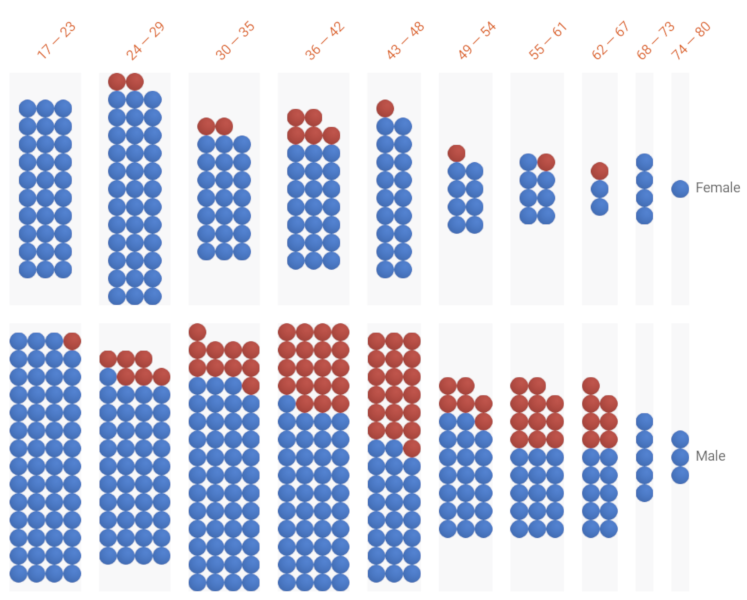}
  \caption{Two-dimensional histogram of age and sex, colored by classification}
  \label{fig:2dHistogram}
\end{subfigure}

\begin{subfigure}{.33\textwidth}
  \captionsetup{width=0.8\textwidth}
  \vspace*{4mm}
  \centering 
  \includegraphics[width=0.9\linewidth]{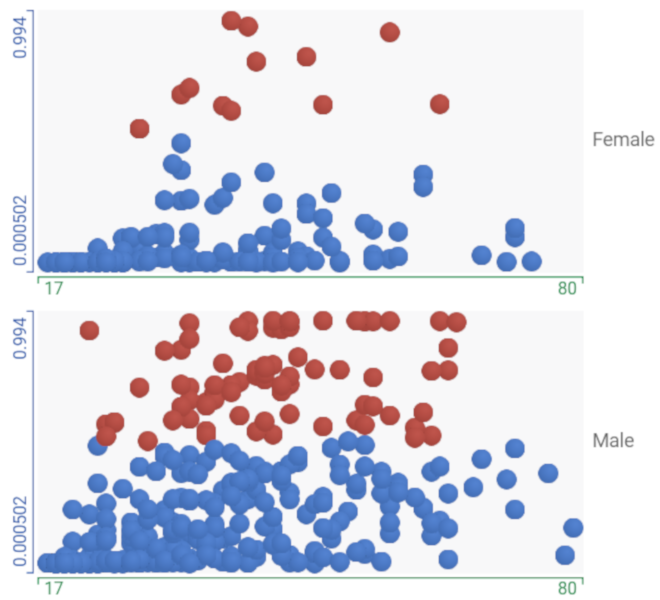}
  \vspace*{2mm}
  \caption{Small multiples by sex. Each scatterplot shows age vs positive classification score, colored by classification}
  \label{fig:smallMultiples}
\end{subfigure}
\begin{subfigure}{.33\textwidth}
  \captionsetup{width=0.8\textwidth}
  \centering 
  \includegraphics[width=0.9\linewidth]{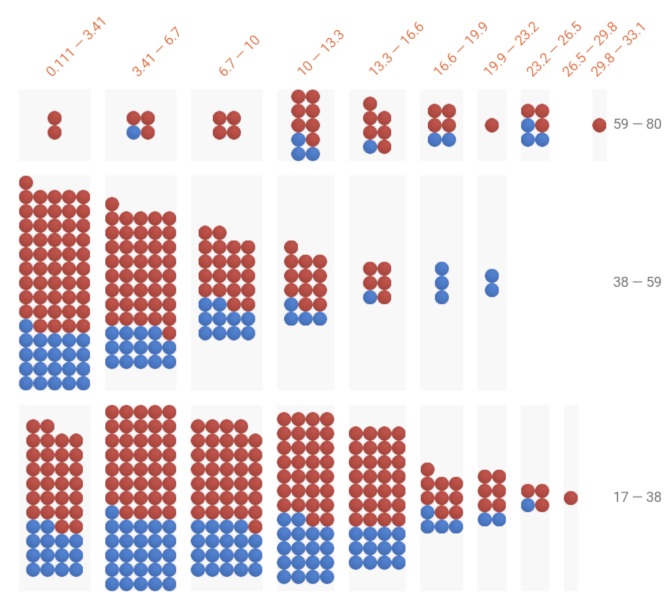}
  \vspace*{6mm}
  \caption{Histograms of performance in a regression model that predicts age, faceted into 3 age buckets}
  \label{fig:manifoldStyle}
\end{subfigure}
\begin{subfigure}{.33\textwidth}
  \captionsetup{width=0.8\textwidth}
  \centering
  \includegraphics[width=0.9\linewidth]{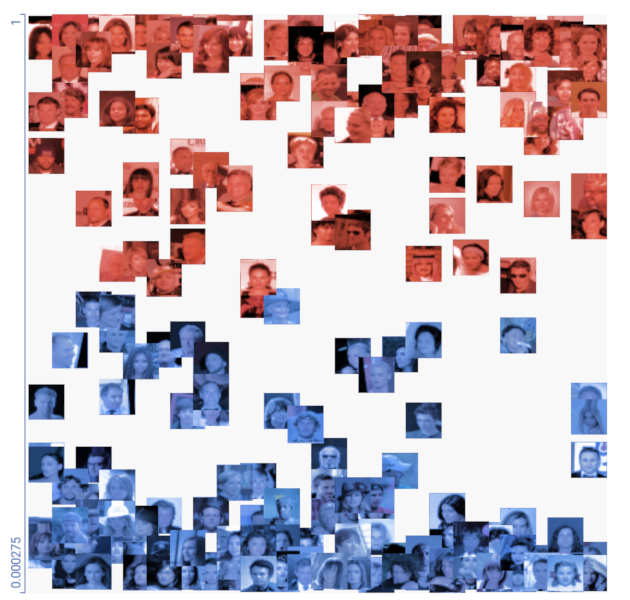}
  \caption{Using images as thumbnails for image datasets}
  \label{fig:imageDive}
\end{subfigure}
\caption{Analyzing two different data sets in the What-If Tool: US Census Income dataset from UCI (a) to (e), and CelebA dataset (f). Three machine-learning tasks are represented: income prediction (a) to (d), age prediction (e), and smile prediction (f).}
\label{fig:variousDivePlots}
}

\begin{document}


\firstsection{Introduction}

\maketitle


People working with machine learning (ML) often need to inspect and analyze the models they build or use. Understanding when a model performs well or poorly, as well as how perturbations to inputs affect the output, are key steps in improving it.
A practitioner may want to ask questions such as: \textit{How would changes to a data point affect my model's prediction? Does my model perform differently for various groups---for example, historically marginalized people? How diverse is the dataset I am testing my model on? }

We present the What-If Tool (WIT), an open-source, model-agnostic interactive visual tool for model understanding. Running the tool requires only trained models and a sample dataset.
The tool is available as part of TensorBoard, the front-end for the TensorFlow \cite{tensorflow} machine learning framework; it can also be used in Jupyter and Colaboratory notebooks.

WIT supports iterative ``what-if'' exploration through a visual interface. Users can perform counterfactual reasoning, investigate decision boundaries, and explore how general changes to data points affect predictions, simulating different realities to see how a model behaves.
It supports flexible visualizations of input data and of model performance, letting the user easily switch between different views.

An important motivation for these visualizations is to enable easy slicing of the data by a combination of different features from the dataset.
We refer to this as  \textit{intersectional analysis}, a term we chose in reference to Crenshaw's concept of ``intersectionality'' \cite{crenshaw2018demarginalizing}. This type of analysis is critical for understanding issues surrounding model fairness investigations \cite{pmlr-v81-buolamwini18a}.

The tool supports both local (analyzing the decision on a single data point) and global (understanding model behavior across an entire dataset) model understanding tasks \cite{DoshiKim2017Interpretability}. It also supports a variety of data and model types.

This paper describes the design of the tool, and walks through a set of scenarios showing how it has been applied in practice to analyze ML systems. Users were able to discover surprising facts about real-world systems--often systems that they had previously analyzed with other tools. Our results suggest that supporting  exploration of hypotheticals is a powerful way to understand the behavior of ML systems.

\section{Related Work}

WIT relates directly to two different areas of research: ML model understanding frameworks and flexible visualization platforms.

\subsection{Model understanding frameworks}

The challenge of understanding ML performance has inspired many systems (see \cite{hohman2018} for a sample survey, or  \cite{patel2008investigating} for a general analysis of the space). Often the goal has been to illuminate the internal workings of a model. This line of work has been especially common in relation to deep neural networks, whose workings remain somewhat mysterious, e.g. Torralba's DrawNet \cite{drawnet} or Strobelt's LSTMVis \cite{DBLP:journals/corr/StrobeltGHPR16}. 

WIT, in contrast, falls into the category of ``black box'' tools; that is, it does not rely on the internals of a model, but is designed to let users probe only inputs and outputs. This limitation reflects real-world constraints (it is not always possible to access model internals), but also means that the tool is extremely general in application.

Several other systems have taken a black-box approach. Uber's Manifold \cite{Zhang_2019}, a tool that has been used for production ML, offers some of the same functionality as WIT. For instance, it features a sophisticated set of visualizations that allow comparison between two models. While Manifold has interesting preset visualizations, our tool implements a flexible framework that can be configured to reproduce the same kind of display as well as supporting many other arrangements through different configurations.
Many other systems are highly specialized, focusing on just one aspect of model understanding. EnsembleMatrix \cite{2009-ensemblematrix} is designed for comparing models that make up ensembles, while RuleMatrix \cite{DBLP:journals/corr/abs-1807-06228} attempts to explain ML models in terms of simple rules.

One of the systems closest in spirit to WIT is ModelTracker \cite{modeltracker}, which provides a rich visualization of model behavior on sample data, and allows users easy access to key performance metrics.
Like our tool, ModelTracker surfaces discrepancies between classification results on similar points.
WIT differs from ModelTracker in several respects.
Most importantly, our tool has a strong focus on testing hypothetical outcomes, intersectional analysis \cite{crenshaw2018demarginalizing}, and machine learning fairness issues. It is also usable as a standalone tool which can be applied to third-party systems.

Another similar system is Prospector\cite{prospector}, a tool that enables visual inspection of black-box models by providing interactive partial dependence diagnostics.
Like WIT, Prospector allows the user to drill down into specific data points to manipulate feature values to determine their effect on model predictions.
But while Prospector relies on the orthogonality of input features for single-class predictive models, WIT affords intersectional analysis of multiple, potentially correlated or confounded features.

Other similar systems include another from Microsoft, GAMut \cite{gamut-a-design-probe-to-understand-howdata-scientists-understand-machine-learning-models}, which allows rich testing of hypotheticals. By design, however, GAMut aims only at the analysis of a particular type of machine learning algorithm (generalized additive models). Another system, iForest \cite{zhaoiforest}, has similar capabilities as WIT, but is solely for use with random forest models.

One significant motivation for our tool comes from the desire to generalize prior visualization work on model understanding and fairness by Hardt, Vi\'egas \& Wattenberg \cite{hardt2016equality}.
A key feature of WIT is that it can calculate ML fairness metrics on trained models. Many current tools offer a similar capability: IBM AI Fairness 360 \cite{aif360-oct-2018}, Audit AI \cite{auditai}, and GAMut \cite{gamut-a-design-probe-to-understand-howdata-scientists-understand-machine-learning-models}. 
A distinguishing element of our tool is its ability to interactively apply optimization procedures to make post-training classification threshold adjustments to improve those metrics. 

\subsection {Flexible visualization platform}

A key element of WIT, used in the Datapoint Editor tab described in more detail below, is the Facets Dive \cite{facets} visualization. This component allows users to create custom views of input data and model results, with an emphasis on exploring the intersection of multiple attributes.

The Facets Dive visualization is related to several previous systems. Its ability to rapidly switch between encodings for X-axis, Y-axis, and color may be a viewed as a simplified version of some of the functionality in Tableau's offering \cite{tableau}. It also relates to Microsoft's Pivot tool \cite{pivot}, both for this same ability and the use of smooth animations to help users understand the transitions between encodings.

Unlike these purely generalist tools, however, Facets Dive is tightly integrated into WIT, and designed for data exploration related to ML. These design choices range from details (e.g., picking defaults that are likely to be helpful to typical ML practitioners) to global architectural decisions. Most importantly, Facets Dive is based entirely on local, in-memory storage and calculation. This protects sensitive data (which may be access restricted, so that sending it to a remote server for analysis is impossible--a common issue with ML data sets). It also allows smooth exploration. at the cost of placing a practical limit on data size.

\section{Background and Overall Design}

We built WIT to empower a larger set of users to probe ML systems, allowing non ML-experts to engage with the technology in an effective way.
At first, we had a fairly broad audience in mind, including data journalists, activists, and civil society groups among others.
However, after testing the tools with a variety of users, it became clear that given some of the technical prerequisites (having a trained model and a dataset), our audience ended up being more technical.
In practice, we restricted ourselves to those with some existing ML experience such as CS students, data scientists, product managers, and ML practitioners. 

Initially, we built a simple proof-of-concept application in which users could interact with a dataset and a trained model to edit datapoints and visualize inference results on the TensorBoard platform. 
Over the course of 15 months, we conducted internal and external studies and designed WIT based on the feedback we received. 
Our internal study recruited four teams and eight individuals within our company who were already in the process of building TensorFlow models and utilizing TensorBoard to monitor them. Participants' use cases represented a variety of example formats (numerical, image-based and text-based), model types (classification and regression) and tasks (predictions and ranking). 
This evaluation consisted of three parts. First, we evaluated the setup of the proof-of-concept application with each participant's model and data. Here, we provided participants with a set of instructions and evaluated the technical suitability and compatibility of model and examples. 
Once participants had a working proof-of-concept with their own model and data, we conducted a think-aloud evaluation in which participants freely explored their model's performance using the system. 
Finally, participants interacted with the proof-of-concept over a period of two to three weeks in their daily jobs, following which we conducted a semi-structured interview. 
Feedback from each studies was incorporated before the next usability study. 

Once WIT was open sourced, we ran an external workshop with five novice machine learning users who provided their own data that they wished to model and evaluate. We introduced participants to WIT through a discussion on model understanding and fairness explorations using interactive demos. We then observed how participants used WIT in the context of their models and data. We concluded the workshop with a discussion reflecting on their use of WIT and a post-workshop offline survey to capture their overall expectations and experience.

A second external WIT workshop was held at the Institute for Applied Computational Science at Harvard University. This workshop was open to the public and was attended by over 70 members of the local data science and machine learning community to maximize diversity of the participant pool. Following a brief introduction to WIT, participants were pointed to demos that they could run using Colaboratory notebooks, along with prompts for investigations. Participants queried the WIT team on the capabilities available and any confusion they faced. Following the workshop, participants were sent an optional offline survey.

The application, as described in this paper, benefits from these multiple rounds of user feedback.

\subsection{User Needs} \label{requirementAnalysis}
Usability studies conducted with early versions of WIT identified a variety of requirements and desired features, which we distilled into five user needs:

\textbf{N1. Test multiple hypotheses with minimal code.} \label{N1}
Users should be able to interact with a trained model through a graphical interface instead of having to write custom code. The interface should support rapid iteration and exploration of multiple hypotheses. This user need corroborates results from  \cite{patel2008investigating}, which suggests that tools for ML practitioners should support (1) an iterative and exploratory process; (2) comprehension of the relationships between data and models; and (3) evaluation of model performance in the context of an application. In the same vein, while visualizing model internals would be useful for hypothetical testing, this would require users to write model-specific code which was generally less favored during usability studies.

\textbf{N2. Use visualizations as a medium for model understanding.}\label{N2}
Understanding a model can depend on a user's ability to generate explanations for model behavior at an instance-, feature- and subgroup-level. Despite attempts to use visualizations for generating these explanations via rules \cite{DBLP:journals/corr/abs-1807-06228}, analogies, and layer activations \cite{cnnvis}, most approaches run into issues of visual complexity, the need to support multiple exploratory workflows, and reliance on simple, interpretable data for meaningful insights \cite{krause2017workflow, cnnvis, DBLP:journals/corr/abs-1807-06228, Zhang_2019}.
WIT should be able to provide multiple, complementary visualizations through which users can arrive at and validate generalized explanations for model behavior locally and globally.

\textbf{N3. Test hypotheticals without having access to the inner workings of a model.} \label{N3}
WIT should treat models as black boxes to help users generate explanations for end-to-end model behavior using hypotheticals to answer questions such as \textit{``How would increasing the value of age affect a model's prediction scores?''} or \textit{``What would need to change in the data point for a different outcome?''}.
Hypotheticals allow users to test model performance on perturbations of data points along one or more specified dimensions.
Without access to model internals, explanations generated using hypotheticals remain model-agnostic and generalize to perturbations.
This increases explanation and representation flexibility, particularly when the models being evaluated are very complex and therefore cannot be meaningfully interpretable \cite{Ribeiro:2016:WIT:2939672.2939778}. 
Hypotheticals are particularly potent when testing conditions of model inference results for new conjectures and what-if scenarios. 

\textbf{N4. Conduct exploratory intersectional analysis of model performance.} \label{N4}
Users are often interested in subsets of data on which models perform unexpectedly. 
For example, Buolamwini and Gebru \cite{pmlr-v81-buolamwini18a} examined accuracy on intersectional subgroups instead of features alone and revealed that three commercially available image classifiers had the poorest performance on the darker female subgroup in test data.
Even if a model achieves high accuracy on a subgroup, the false positive and false negative rates can be wildly different, leading to real-world consequences\cite{compas}.
Since there are multiple ways to define these subsets, visualizations for exploratory data analysis should retain flexiblity, scalablity and customization for input data types, model task and optimization strategies. 
Complementary views at instance-, feature-, subset- and outcome- levels are particularly insightful when comparing the performance of multiple models on the same dataset. 

\textbf{N5. Evaluate potential performance improvements for multiple models.} \label{N5}
In model development, it can be hard to track the impact of changes, such as changing a classification threshold.
For example, optimizing thresholds for demographic parity may ensure that all groups receive an equal fraction of ``advantaged'' classification, while putting sub-groups with lower true positive rates at a disadvantage \cite{hardt2016equality}. 
One may want to test different optimization strategies for a variety of costs locally and globally before changing the training data or model hyperparameters to improve performance.
Users should be able to interactively debug model performance by testing strategies that mitigate undesirable model behaviors in an exploratory space on smaller datasets without needing to take the analysis offline. 
A closely related way that users test changes to a model is by comparing the performance of a \textit{``control''} model to \textit{``experimental''} versions on benchmarks.

\subsection {Overall Design} \label{visualizationInterface}
WIT is part of the TensorBoard application \cite{tensorboard}, as a code-free, installation-free dashboard. 
It is also available as a standalone notebook extension for Jupyter and Colaboratory notebooks.

In TensorBoard mode, users can configure the tool through a dialog box to load a dataset from disk and to query a model being served through TensorFlow Serving. In notebook mode, setup options are provided through the Python invocation that displays the tool. 

In addition to being useful for model understanding efforts on a single model, WIT can be used to compare results across two models for the same dataset. In this case, all standard features apply to both models for meaningful comparisons.

Out-of-the-box support, with no need for any custom coding by the user in alignment with user need \textbf{N1}, is provided for those making use of the TensorFlow Extended (TFX) pipeline \cite{tfx}, through WIT's support of TFRecord files, TensorFlow Estimator models, and model serving through TensorFlow Serving. In notebook mode, the tool can be used with models and datasets outside of this paradigm, through use of a user-provided Python function for performing model prediction.

WIT's interface consists of two main panels: a visualization panel on the right and a control panel on the left. The control panel has three tabs designed to support exploratory workflows: Datapoint Editor, Performance + Fairness, and Features. 

\section{Task-Based Functionality} \label{functionality}

Here we discuss the implementation of WIT, and how it builds on user needs outlined in \ref{requirementAnalysis}, in the context of different user tasks. 

\begin{figure*}[tbh]
 \centering
 \fbox{\includegraphics[width=\textwidth]{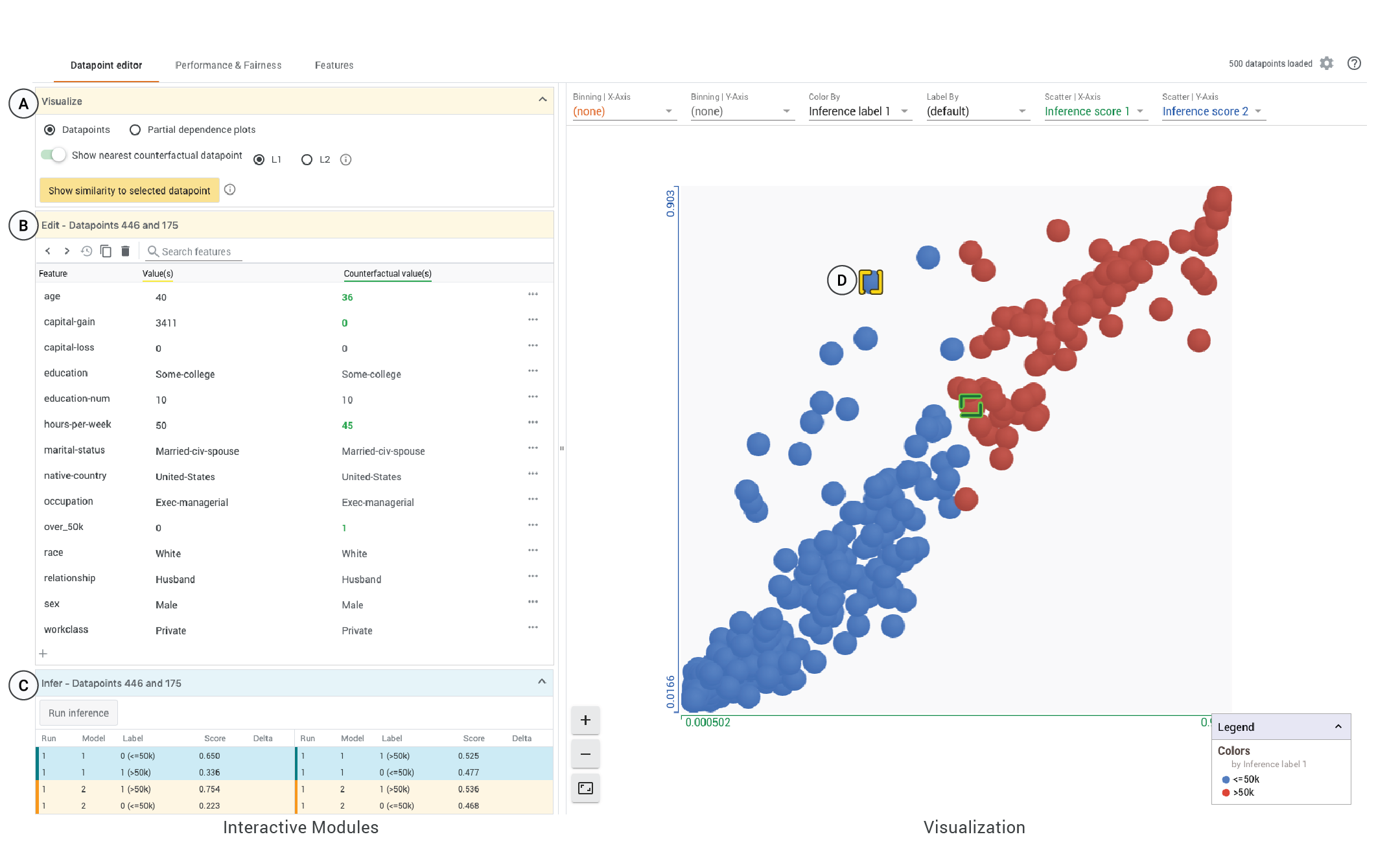}}
  \caption{Datapoint Editor in WIT. The left panel displays information related to machine-learning aspects of a selected data point in the dataset, including a list of feature values (B), inference values (C) and counterfactual controls (A). The right panel visualizes the dataset and offers controls for visually slicing the data according to its features. In (B) users can edit values of the data point and re-run inference to examine results. (B) also shows the closest counterfactual value to the selected data point, highlighting the delta between the two points. (D) is the selected data point.
  }
  \label{fig:teaser}
\end{figure*}

We utilize the UCI Census dataset \cite{Dua2019} throughout this section as a running example to describe and contextualize WIT's features.
We analyze two models trained on the UCI dataset and prediction task \cite{Dua2019} which classify individuals as high ($\geq$\$50K/year) or low income from their census data. 
\textit{Model 1} is a multi-layer neural network and \textit{Model 2} is a simple linear classifier.
We test both models on 500 data points from the UCI test dataset\footnote{This particular demo, along with a link to the tool's source code, can be found on the What-If Tool website (https://pair-code.github.io/what-if-tool/).}. 
In particular, we examine a selected data point, \textbf{D} (highlighted in yellow in Figure \ref{fig:teaser}), where the two models disagree on the prediction.
We explore how the \textit{capital-gain} feature disparately influences the prediction for the two models.

\subsection{Exploring Your Data}
\label{exploringYourData}

In this section we describe how users can explore their data as well as perform customizable analyses of their data and model results.

\subsubsection{Customizable Analysis}\label{customizableAnalysis}

The Datapoint Editor tab of WIT shows all loaded data points and their model inference values. 
Data points can be binned, positioned, colored and labeled by values of any of their features. 
Each data point is represented by a circle or a thumbnail image.
The visualization allows users to zoom and pan around the data points, while describing details such as the categorical color scheme in a legend.

To support interactive evaluations of model results on intersectional subgroups, users can create visualizations using performance metrics from the model, in addition to combinations of the features in the dataset and model inference values.
With respect to performance metrics, in classification models users can color, label or bin data points by the correctness of a classification.
In regression models, users can color, label or bin data points by inference error, inference mean error, or inference absolute error, plus position data points in a scatterplot by those same error measures.


The Datapoint Editor enables users to create custom visualizations from both the dataset and model results for a variety of different model understanding tasks. Some useful visualizations include:
    \begin{itemize}
        \item \textbf{Confusion matrices} [Figure \ref{fig:confMatrix}] for binary and multi-class classifiers. In the figure, binning is set to the ground truth on the X-axis and model prediction on the Y-axis. Datapoints are colored by correctness of the prediction, where \textcolor{blue}{blue} is correct and \textcolor{red}{red} is incorrect.
        \item \textbf{Histograms and bar or column charts} [Figure \ref{fig:ageHistogram}] for categorical and numeric features. For numeric features, binning is done with uniform-width bins and the number of bins is adjustable, defaulting to 10. In this figure, datapoints are colored by classification by model 1, where datapoints colored \textcolor{blue}{blue} are positively classified and those colored \textcolor{red}{red} are negatively classified. 
        \item \textbf{2-dimensional histograms/bar/column charts} [Figure \ref{fig:2dHistogram}] with binning on the X-axis and Y-axis by separate features.
        \item \textbf{2D histogram of prediction error} [Figure \ref{fig:manifoldStyle}] of data points, mimicking performance visualizations used in the Manifold system \cite{Zhang_2019}.  Here, datapoints are colored by sex where \textcolor{red}{red} is male and \textcolor{blue}{blue} is female.
        \item \textbf{Small multiples of scatterplots} [Figure \ref{fig:smallMultiples}], created by binning by one feature or the intersection of two features, and creating scatterplots within those bins using numeric features.
    \end{itemize}

In our Census example, since we compare two binary classification models, the visualization defaults to a scatterplot that compares scores for the positive classes for the linear classifier and the deep neural network on the Y- and X- axis respectively (See Figure \ref{fig:teaser}).
Data points that fall along a diagonal represent individuals for whom the models tend to agree.
Those further away from the diagonal indicate disagreement between the two models.

\subsubsection{Features Analysis: Dataset Summary Statistics}

\begin{figure*}[ht]
 \center
  \includegraphics[width=\textwidth]{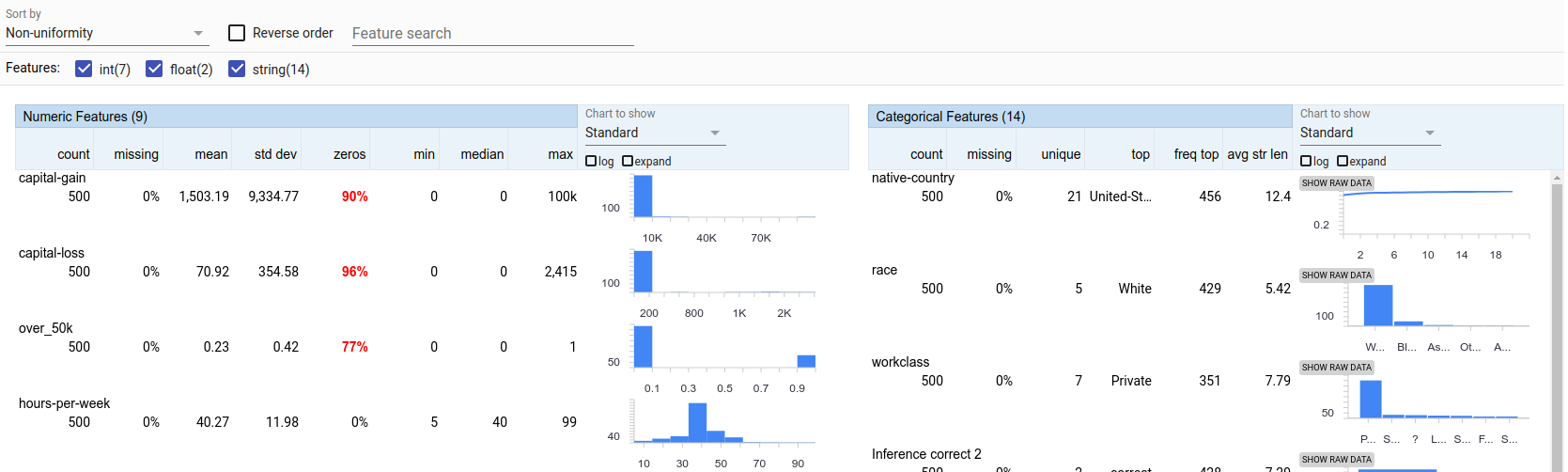}
  \caption{Features tab in WIT, showing summary statistics for the UCI Census dataset. Features are sorted by non-uniformity of the distributions of their values. Capital gains and capital loss have a large percentage of zeros and a long tail of non-zero values.}
  \label{fig:overview}
\end{figure*}

Understanding the data used to train and test a model is critical to understanding the model itself. 
The Features tab of WIT builds on the Facets Overview \cite{facets} visualization to automatically provide users with summary statistics and charts of distributions of all features in the loaded dataset.
A user may find these statistics and visualizations particularly useful when validating explanations originating from model performance on subsets of the loaded dataset, in line with user need \textbf{N2}. For example, the NIST gender report \cite{NIST} evaluates the impact of ethnicity on gender classification, using country of origin as a proxy for ethnicity.
However, Africa and the Caribbean, which have significant Black populations are not represented in the 10 locations used in the report.
By visualizing the distributions of features, users can gain a sense for unique feature values which may not be discernible through dataset metadata alone.

For numeric features, minimum, maximum, mean, and standard deviation are displayed along with a 10-bin equal width histogram describing the distribution of feature values.
For categorical features, statistics such as number of unique values and most frequent value are displayed along with a visualization of the distribution of these values.
Users can toggle between reading a distribution as a chart or as a table.
Features with few distinct values ($\leq$20) are displayed as histograms; for others, to avoid clutter, WIT shows a cumulative distribution function as a line chart.
Exact values are provided on tooltips, and users can control the size of the charts and switch between linear or log scale.

WIT offers multiple ways to sort features. For instance, a user may sort by non-uniformity to see features that have severely unbalanced distributions, or sort by number of zeros or missing values to see features that have abnormally large counts of empty feature values. 

For our Census example, Figure \ref{fig:overview} shows a huge imbalance in the distribution of values for capital gains, with 90\% of the values being 0, but having a small set of values ranging all the way up to 100,000.
This helps explain the results in section \ref{investigatingWhatIfHypotheses}, where the neural network model learns to essentially ignore zero values for capital gains.

\subsection{Investigating What-If Hypotheses}
\label{investigatingWhatIfHypotheses}

Here we show how users can generate and test hypotheses about how their model treats data by identifying counterfactuals and observing partial dependence plots.

\subsubsection{Data Point Editing}
\label{dataEditing}

While it has been shown that in some circumstances providing model explanations (answering "Why?" or "Why Not?" questions) can increase user performance and trust\cite{lim_2009},
we've found--like others\cite{prospector}--that one of the most powerful ways to analyze a model is in terms of \textit{hypotheticals}: probing how an output would change based on carefully chosen input modifications (see section \ref{mitStudentProject} for one case study).
For example, in the context of a model making bank loan determinations, one might ask questions such as, ``Would person X have gotten a loan if she were male instead of female?'' or ``How much would a small increase in person X's income have affected the result?''

WIT makes it easy to edit data and see the effect on the model's inferences.
The user can select a data point in the Datapoint Editor tab and view its feature values and results of model predictions. WIT shows prediction scores for each available model, either as regression scores or as class scores for the top returned classes in classification models. 

\begin{figure}[ht]
 \centering
 \includegraphics[width=\linewidth]{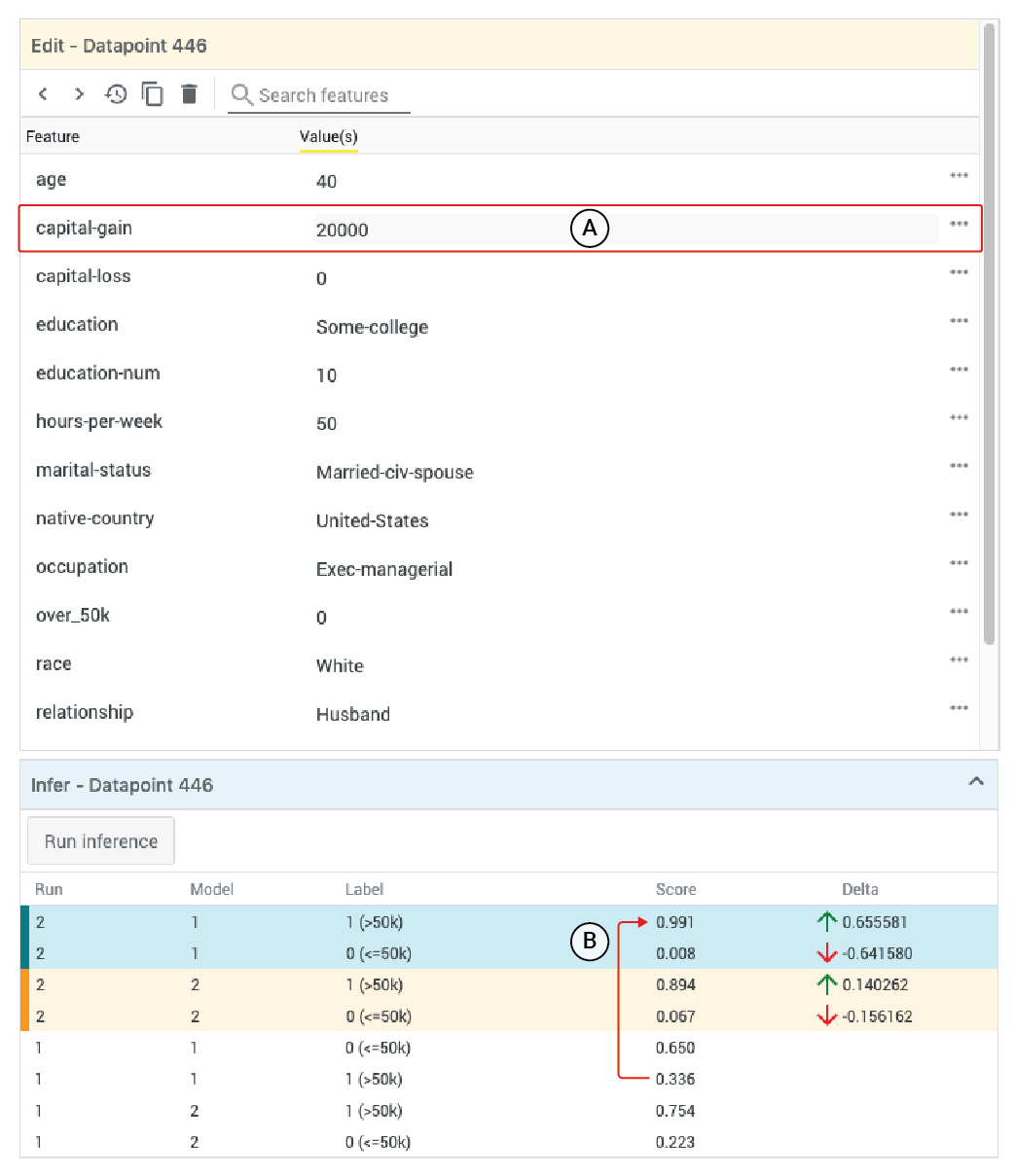}
  \caption{Datapoint editor view showing an edited value for “capital-gain” (A), which the user has changed from 3,411 to 20,000. After rerunning inference (B), this edit causes the score for label 1 ($\geq$50K) to change from 0.336 to 0.991, essentially flipping the prediction for this datapoint.}
  \label{fig:datapointEditing}
\end{figure}

To conduct iterative what-if analyses, the user can edit, add, or delete individual feature values or entire features within an instance in the editor module and see the effect those changes have on the model prediction for that datapoint.
Additionally, new features can be added to the datapoints solely for visualization purposes; for example, tagging a set of examples as "high priority" and then visualizing separate confusion matrices for normal and "high priority" datapoints.
Image-type features can be edited by replacing existing images. Edited data points are then re-inferred by the model and the inference module is updated with new scores, as well as the delta and direction of the change. 
Users can duplicate or delete selected data points, useful when comparing multiple small changes in a single data point to the original. 
For our Census example, Figure \ref{fig:datapointEditing} shows editing values for the yellow-bordered, selected data point from Figure \ref{fig:teaser}. This data point initially has opposite classifications from the two models. Editing this data point to increase its capital gains and re-inferring shows that the neural network model then changes the prediction to the positive class, so it now matches the prediction from the linear model, which predicted the positive class both before and after the edit. We will continue to explore the effect of changing capital gains on this data point in the sections \ref{counterfactuals} and \ref{pdp}.

\subsubsection{Counterfactual Reasoning}\label{counterfactuals}

It can also be helpful to explore an inverse type of hypothetical question.
Consider, for the bank loan example, that for a given person the model predicts that the loan would be denied.
A user of WIT who is interrogating this model might ask, \textit{``What would have to change so that the person X would get the loan?''}. 
In cases like this there are often many possible answers, but typically practitioners are most interested in comparing differences to data points upon which the model predicted a different outcome.
In the loan example, this means answering the question, \textit{``Who is the person most similar to person X who got a loan?''}
 
Such data points have been termed \textit{counterfactual examples} by Wachter \cite{wachter2017counterfactual} (Note that this terminology can cause some confusion, since it is not identical in usage to that of Judea Pearl and collaborators \cite{pearl2014probabilistic}).
Our tool provides users a way of automatically identifying such counterfactual examples for any selected data point.

To sort possible counterfactual examples for a given selected data point, without the need for the user to provide their own code or specification, the tool uses a simple distance metric which aggregates the differences between data points' feature values across all input features.
To do this, it treats numeric and categorical feature values slightly differently to generate distances between data points:
    \begin{itemize}
        \item \textbf{Numeric features:}
Absolute value of the difference between the two data points divided by the standard deviation of that feature across the entire dataset.
        \item \textbf{Categorical features:}
Data points with the same value have a distance of 0, and other distances are set to the probability that any two examples across the entire dataset would share the same value for that feature.
    \end{itemize}
To compute the total distance between two data points, the tool individually assesses the distance for each numeric and categorical feature, and then aggregates these using either an $L^1$ or $L^2$ norm, as chosen by the user. The default measure uses $L^1$ norm as it is a more intuitive measure to non-expert users. We provide the option to use $L^2$ norm as early users requested this flexibility to help them further explore possible counterfactuals.

For our Census example, Figure \ref{fig:teaser} shows the currently selected, negatively classified data point (blue dot with yellow border), and the most similar, positively classified, counterfactual data point (red dot with green border), for classifications from the neural network model.
There are only a few differences between the feature values of the selected data point and its counterfactual, shown in the left-hand panel of Figure \ref{fig:teaser}, highlighted in green.
Specifically, the counterfactual has a slightly lower age, higher hours worked per week, and a reported capital gain of \$0 (as opposed to \$3,411). 
That last value might seem surprising, as one might expect that having a significantly lower value for capital gains would indicate a lower income (negative classification) than non-zero capital gains would. Especially considering that in section \ref{dataEditing}, it was shown that raising the capital gains for the selected data point would also cause the neural network model to change its classification to the positive class.

In addition to inspecting the most similar counterfactual example for a given selected data point, the user can elevate the $L^1$ or $L^2$ distance measure to become a feature in its own right.
Such a distance feature can then be used in custom visualizations for more nuanced counterfactual reasoning, such as a scatterplot in which one axis represents the similarity to a selected data point.

\subsubsection{Partial Dependence Plots}\label{pdp}
Often practitioners have questions about the effect of a feature across an entire range of values. To address these questions, the tool offers partial dependence plots, which show how model predictions change as the value of a specific feature is adjusted for a given data point. 

The partial dependence plots are line charts for numeric features and column charts for categorical features. In the numeric case, we visualize class scores (or regression scores for regression models) for the positive class (or top-N specified classes, when there are more than two classes) at 10 equally distributed points between the minimum and maximum observed values for a feature. Users can also manually specify custom ranges.
For categorical features, class scores or regression scores are displayed as column or multi-series column charts, where the X-axis consists of the most common values of the given feature across the loaded dataset. 

No partial dependence plots are generated for images or for features with unique values (typically these represent something like IDs, so the chart would be difficult to read and not meaningful).
When comparing multiple models, inference results for both models are shown on the same plot.
In each plot, we indicate original feature values and classification thresholds for all models wherever applicable. 
Users can hover for tooltips that contain legends and exact values along the plots.

WIT also provides instance-agnostic \textit{global }partial dependence plots, which are computed by averaging inference results on all data points in the dataset. 
This is particularly useful when checking if the relationship between a feature and model's performance is consistent locally and globally.
\begin{figure}
 \centering
 \includegraphics[width=\linewidth]{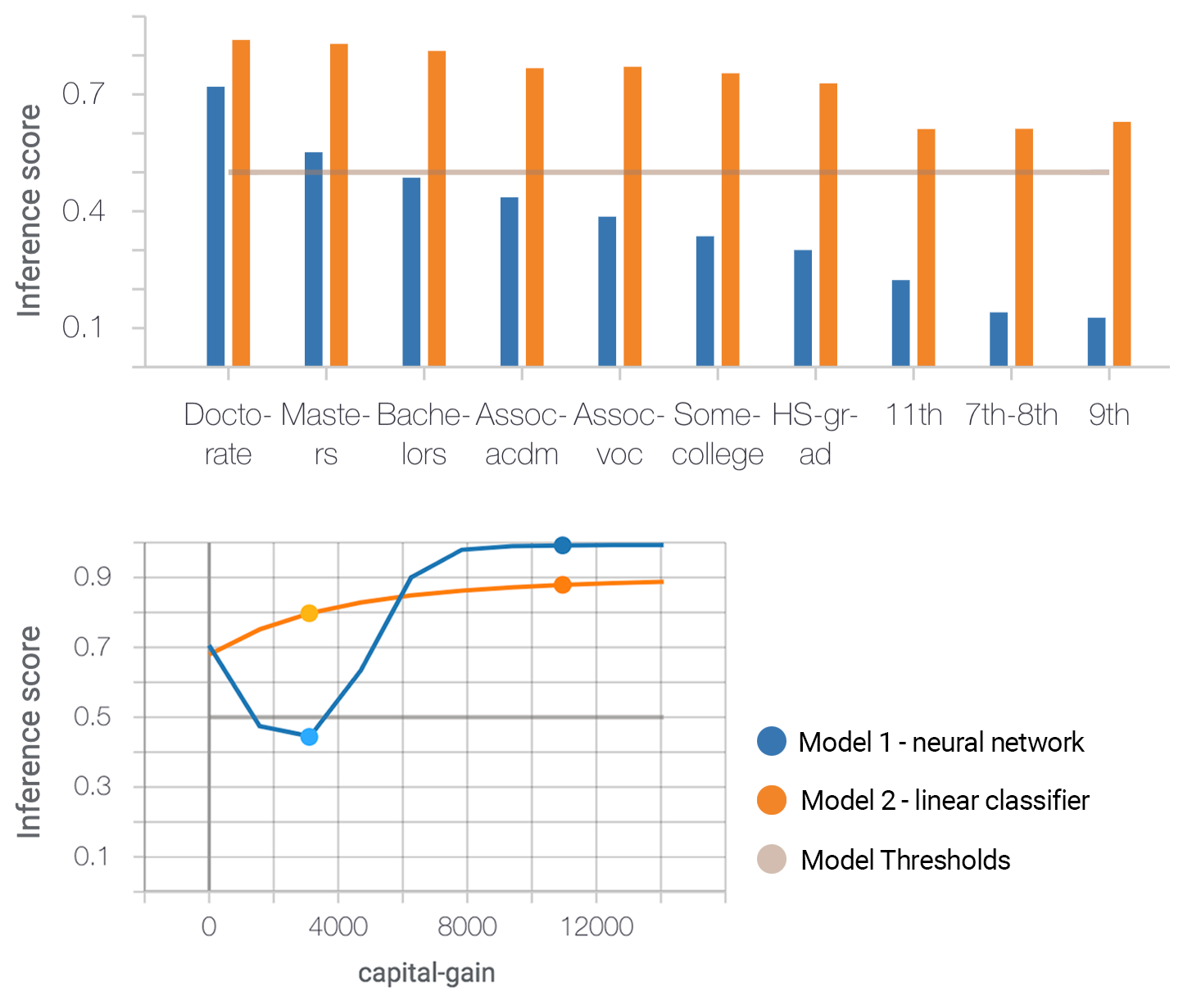}
  \caption{Partial Dependence (PD) plot view. Top: categorical PD plot for “education level.” Bottom: numeric PD plot for “capital gains” feature. The neural network has learned that low but non-zero capital gains are indicative of lower income, high capital gains are indicative of high income, but zero capital gains is not necessarily indicative of either. In both plots, the horizontal gray line represents the positive classification threshold for the models.}
  \label{fig:pd}
\end{figure}

For our Census example, Figure \ref{fig:pd} depicts partial dependence plots for the \textit{education} and \textit{capital-gain} features. We find that the inference score and capital-gain feature have a monotonic relationship in the linear model (as expected), whereas the neural network has learned a more nuanced relationship, where both having zero capital gains and having higher capital gains are indicative of high income rather than having a small amount of capital gains. This is consistent with the observed counterfactual behavior, discussed in \ref{counterfactuals}.

\subsection{Evaluating Performance and Fairness}
\label{evaluatingPerformanceAndFairness}

To meet needs \textbf{N4} and \textbf{N5}, WIT provides tools for analyzing performance, available under a ``Performance + Fairness'' tab. The tools can be used to analyze aggregate model performance as well as compare performance on slices of data. Following the same layout paradigm as the Datapoint Editor tab, users can interact with configuration controls on the left side panel to specify ground truth features, select features to slice by, and set cost ratios for performance measures, and view the resulting performance table on the right side.
An initial view of the performance table offers users performance measures tailored to the type of prediction task.
Users can slice their data into subgroups by a single feature available in the dataset, or by the intersection of two features, in order to enable intersectional performance comparisons. For example, in our UCI Census example, users could slice by both sex and race to view intersectional performance, similar to the type of analysis done in \cite{pmlr-v81-buolamwini18a}.

Slicing by features calculates these performance measures for subgroups, which are defined by the unique values in the selected features.
\subsubsection{Performance Measures}
When applied, slices are initially sorted by their data point count and can also be sorted alphabetically or by performance measures such as accuracy for a classification model or mean error for a regression model.
Table \ref{tab:performanceMeasuresTable} describes the different performance measures available for the supported model types.
\begin{table}[htb]
    \centering
        \begin{tabular}{|r|l|}
         \hline
         \textbf{Model Type} & \textbf{Performance Measures}\\
        \hline
        Binary classification
         & Accuracy \\
         & False positive percentage\\
         & False negative percentage\\
         & Confusion matrix\\
         & ROC curve\\
        \hline
        Multi-class classification
         & Accuracy\\
         & Confusion matrix\\
        \hline
        Regression
         & Mean error\\
         & Mean absolute error \\
         & Mean squared error\\
        \hline
        \end{tabular}
    \caption{Performance measures reported for different model types in the Performance + Fairness tab.}
    \label{tab:performanceMeasuresTable}
\end{table}

For binary classification models, receiver operating characteristic (ROC) curves plot the true positive rate and false positive rate at each possible classification threshold for all models in a single chart. For both binary and multiclass classification models, confusion matrices are provided for each model. Error counts are emphasized graphically via color opacity, as shown in Figure \ref{fig:perf}.

In the rest of this section, we explain details specific to binary classification, such as cost ratios and positive classification threshold values.

\subsubsection{Cost Ratio}
ML systems can make different kinds of mistakes: false positives and false negatives. The cost ratio determines how ``expensive'' these different kinds of errors might be. In a sense, by setting the cost ratio, users decide how conservative their system should be. For example, in a medical context, it might be preferable for a system to rather sensitive, where it is much more likely to give false positives (i.e. a screening test reports cancer when there is no cancer), instead of the other way around (e.g. the screening test misses cancer signs in a patient).

In WIT, users can change the cost ratio, in order to automatically find the optimal positive classification threshold for a model given the results of model inference on the loaded dataset, in line with user needs \textbf{N1} and \textbf{N5}. The positive classification threshold is the minimum score that a binary classification model must give for the positive class before the model labels a data point as being in the positive class, as opposed to the negative class.

When no cost ratio is specified, it defaults to 1.0, at which false positives and false negatives are considered equally undesirable. 
When optimizing the positive classification threshold with this default cost ratio, WIT will find the threshold which achieves the highest accuracy on the loaded dataset. Upon setting a threshold, it displays the performance measures when using that threshold to classify data points as belonging to the positive class or the negative class.

For our Census example, Figure \ref{fig:perf1} shows the performance of the two models, broken down by sex of the people represented by the data points, with the positive classification thresholds set to their default values of 0.5. It can be seen that the accuracy is higher for women, and from the first columns of the confusion matrices, that there is a large disparity in the percentage of data points predicted as being in the positive class for men and women. This disparity also exists in the actual ground truth classes from the test dataset between the two sexes as well, as shown in the first rows of the confusion matrices.

\subsubsection{Thresholds and Fairness Optimization Strategies}
In addition to evaluating metrics such as accuracy and error rates for specified slices in the performance table, users can explore different fairness optimization strategies.
Classification thresholds for each slice can be individually modified so that the positive class threshold is different depending on the data point; the performance table and visualization instantly update to reflect resulting changes in performance.

Applying a fairness optimization strategy through the tool automatically updates the classification thresholds for each slice individually in order to satisfy a particular fairness definition.
Again, the results can be seen across the performance table with updates to thresholds, performance metrics and visualizations.
Table \ref{tab:fairnessDefinitions} describes the fairness optimization strategies for binary classification available in the tool.

\begin{table}[htb]
    \centering
    {\footnotesize
    \begin{tabular}{|p{0.3\columnwidth}|p{0.59\columnwidth}|}
    \hline
         \textbf{Fairness Strategy} & \textbf{Definition}\\
        \hline
        Single Threshold & A single threshold for all data points based solely on the specified cost ratio.\\
        \hline
        Group Thresholds & Separate thresholds for each slice based solely on the specified cost ratio.\\
         \hline
        Demographic Parity & Similar percentages of data points from each slice must be predicted as positive classifications.\\
        \hline
        Equal Opportunity & Similar percentages of correct predictions must exist across each slice for those data points predicted as positive classifications. \cite{hardt2016equality} \\
        \hline
        Equal Accuracy & Similar percentages of correct predictions must be made across each slice.\\
         \hline
    \end{tabular}
    }
    \caption{Fairness optimization strategies available in WIT.}
    \label{tab:fairnessDefinitions}
\end{table}

\begin{figure*}
\begin{subfigure}{.5\textwidth}
  \centering
  \fbox{\includegraphics[width=.9\linewidth]{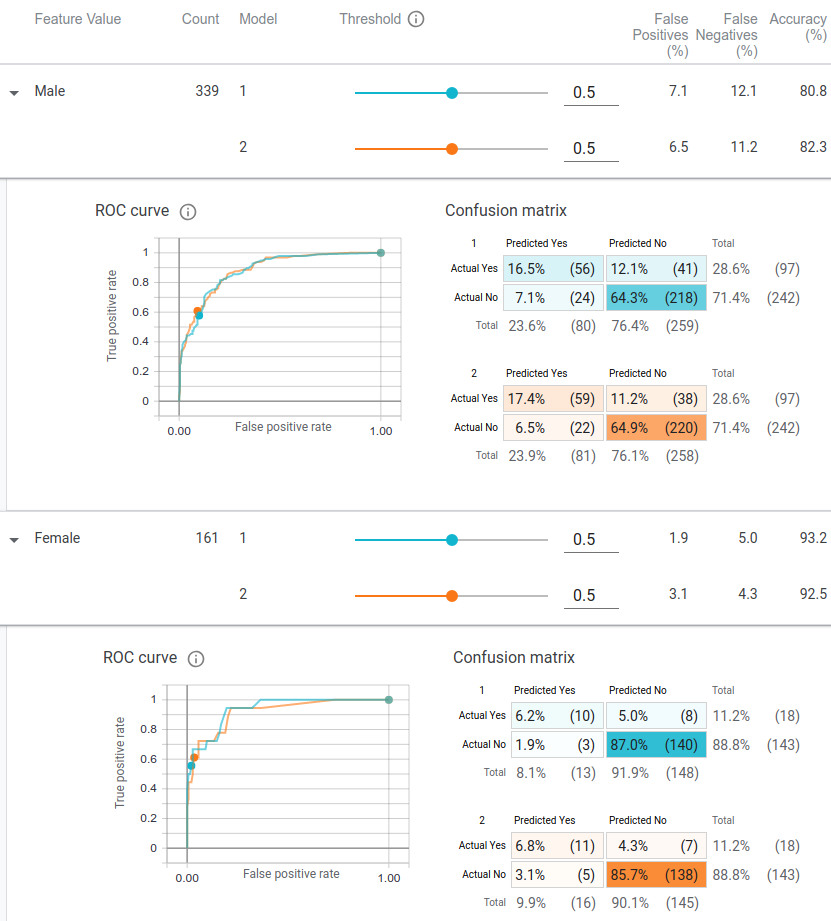}}
  \caption{Default positive classification thresholds.}
  \label{fig:perf1}
\end{subfigure}
\begin{subfigure}{.5\textwidth}
  \centering
  \fbox{\includegraphics[width=.9\linewidth]{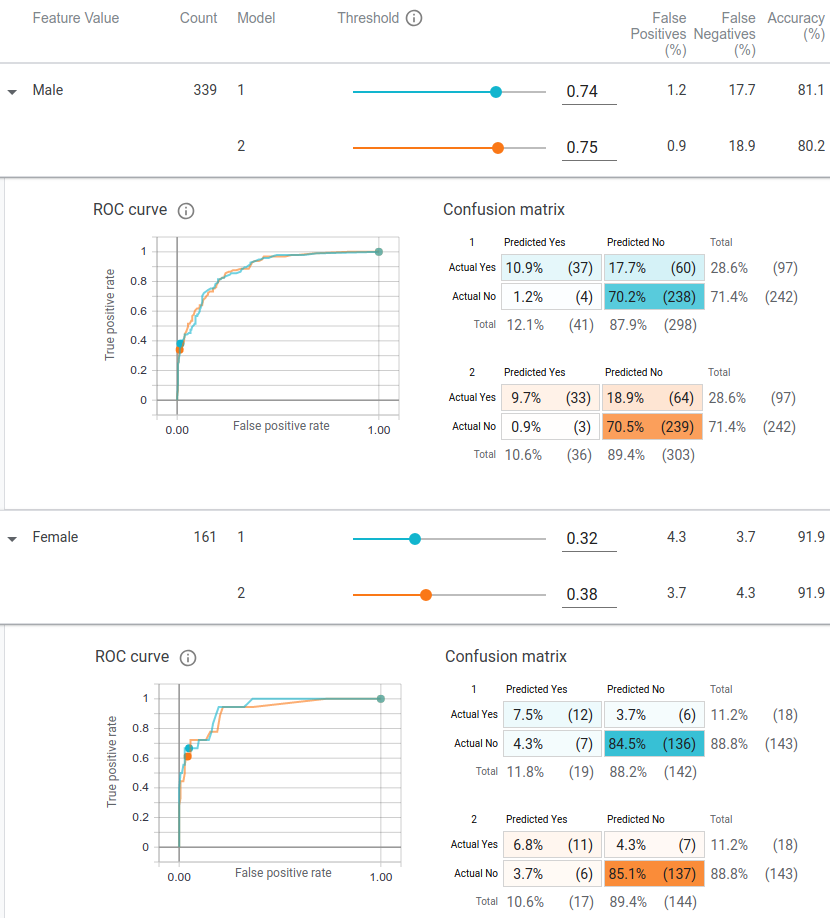}}
  \caption{Positive classification thresholds set for demographic parity between sexes.}
  \label{fig:perf2}
\end{subfigure}
\caption{Performance view of our Census example models broken down by sex, (a) showing performance when the positive classification thresholds are left at their default levels of 0.5 for each sex, and (b) showing performance when the thresholds have been set to achieve demographic parity between sexes. Each confusion matrix refers to a given model: teal for model 1 (neural network) and orange for model 2 (linear classifier). Achieving demographic party on either model requires lowering the threshold for females and raising it for males.}
\label{fig:perf}
\end{figure*}

In our Census example (Figure \ref{fig:perf2}), when optimizing the models' classification thresholds for demographic parity between sexes, the thresholds for male data points is raised and the threshold for female data points is lowered.
This adjustment accounts for the large imbalance in the UCI Census dataset between sexes, where men are much more likely to be labeled as high income than women.

\subsection{Comparing Two Models}
\label{comparingModels}
It is often desirable to see the performance of a model with respect to another one. For instance, a practitioner may make changes to their datasets or models based on the feedback they receive from our tool, train a new model, and compare it to the old one to see if the fairness metrics have improved. To facilitate this workflow, all performance measures and visualizations, such as ROC curves and partial dependence plots, support comparison of two models. For example, when using the tool with two models, it immediately plots an inference score comparison between the models, allowing one to pick individual data points that differ in the scores and analyze what caused this difference. One can use partial dependence plots to compare if the new model is more or less sensitive to a feature of interest than an old model.

\subsection{Data Scaling}
\label{DataScaling}
Here we provide approximate numbers on how many data points can be loaded into WIT for a standard laptop from the last several years.
For tabular data with 10-100 numeric or string inputs (such as the UCI Census dataset), WIT can handle $\sim 100k$ points. For small images (78x64 pixels), WIT can load $\sim 2000$ points.
In general, the number of features and the size of each feature are the main factors for memory constraints. For example, it would be possible to load more images using smaller image sizes.

\section{Case studies}

We present three case studies to show how WIT addresses user needs and can provide insights into ML model performance. The first two cases come from a large technology company. The third involves a set of undergraduate engineering/computer science students analyzing models to gain insights about a particular dataset.

\subsection{Regression Model from an ML Researcher}

An ML researcher at a large technology company used our tool to analyze a regression model used in a production setting.
In a post-use interview, they said they wanted to better understand ``how different features affect the predictions'' (aligning with user need \textbf{N3}).
They loaded a random sample of 2,000 data points from the test dataset and the model used in production.

To understand how different features affect the model's output, the researcher picked individual points from the Datapoint Editor tab, manually edited  feature values, and then re-ran inference to see how regression values changed.
After doing this a number of times, they moved over to partial dependence plots to further understand how altering each feature in a data point would change the model's regression score, validating the tool's ability to satisfy user need \textbf{N2}.
They immediately noticed something odd: the partial dependence plot for a certain feature was completely flat for every data point they looked at. This led to the hypothesis that the model was never taking that feature into account during prediction--a surprise, since they expected this feature to have a significant effect on the final regression score for a data point.

With this new information, they investigated the code in their ML pipeline, and found a bug preventing that feature from being ingested by the model at serving time.
WIT helped them uncover and fix an instance of training-serving skew that was unknowingly affecting their model's performance.
Note that this bug had existed for some time and had not been detected until this researcher used our tool to analyze their model.

\subsection{Model Comparison by a Software Engineer}

A software engineer at a large technology company used WIT to compare two versions of a regression model.
The models predict a specific health metric for a medical patient, given a previous measurement of that metric, the time since that first measurement, and some other features.
The first model was trained on an initial dataset and the second model was trained on a newer dataset which was generated with slightly different feature-building code to create the input features for the model to train and evaluate on.

The engineer used our tool to compare the performance of the two models.
They noticed immediately through the Datapoint Editor visualization and inference results of selected data points that the first model's predictions were almost always larger than the second model's.
They then plotted the target value of each test data point's prediction against the initial health metric measurement across data from both the training set of both models. This exemplifies the tool's capability to satisfy user need \textbf{N2}.
For the first model's training data, the target value was almost always larger than the initial measurement.
They did not expect to see this relationship, and did not see it for data points from the second model's dataset.
An investigation uncovered a bug in the code that builds the input features for the first model, causing it to swap the first and last value of the measurements.
In the engineer's words, ``before using the What-If Tool we didn't suspect any bug because the same feature building code was used for all training/validation/test sets. So the error metrics were reasonable. The model was just solving a different problem.''
The success of the tool in uncovering a previously-unnoticed issue directly relates to user need \textbf{N5}.

\subsection{Binary Classification, Counterfactual and Fairness Metrics Exploration by University Students} \label{mitStudentProject}

A group of MIT students working on a project for the \textit{Foundations of Information Policy} course made extensive use of WIT.
Their goal was to analyze the fairness of the Boston Police Departments stop-and-frisk practices. 
They took the Boston Police Department Field Interrogation and Observation (FIO) Dataset \cite{bpd}, which provides over 150,000 records of stop and frisk encounters from 2011 to 2015, and created a linear classifier model to emulate those decisions using a training set sampled from that dataset. This model could then be queried to reason about the department's decision making process, beyond just looking the raw data.
Their model takes as input information about a police encounter, such as the age, race, gender, and criminal record of an individual being stopped by police, along with the date, time, and location of the encounter and the name of the officer, and as an output determines whether that person was searched or frisked (positive class) or not (negative class) during the encounter.
After training a model to perform this classification, they used WIT to analyze the model on the held-out test set in order to reason about the department's stop-and-frisk practices. By analyzing their model, as opposed to the raw dataset, the students were able to ask a variety of ``what-if'' questions, aligning with user need \textbf{N1}.

The group made extensive use of the Datapoint Editor tab to explore individual data points and perform counterfactual reasoning on them.
They explored data points where the model's positive classification score was close to 0.5, meaning the model was very unsure of which class the data point belonged to.
They further tested these points by altering features about the data point and re-inferring to see how the prediction changed, both with manual edits and through the partial dependence plots.

The students were not originally planning on doing much analysis around the effect of the police officer ID on prediction, but by using the nearest counterfactual comparison tool, they quickly realized that predictions on many data points could have their classification flipped just by changing the officer ID.
This was immediately apparent when seeing that on many data points the closest counterfactual was identical (ignoring encounter date and time) other than the officer ID.

They were able to conclude that although age, gender and race were important characteristics in predicting the likelihood of being frisked, as they had theorized, the most salient input feature was the officer involved in the encounter, which was something they did not anticipate.
This use of WIT illustrates user need \textbf{N3}.

The students were also interested in analyzing their model for equality of opportunity for different slices of the dataset, probing for fairness across different groups.
Using the WIT's Performance + Fairness tab, they sliced the dataset by the features they were interested in analyzing and observed the confusion matrices for the different groups when using the same positive classification threshold.
The students then used different optimization strategies to check how the thresholds and model performance would change when adjusting the thresholds per slice to achieve different fairness constraints, such as equality of opportunity.

They concluded that in their model, certain groups of people were disadvantaged when it came to the chance of being frisked during a police encounter: \textit{``Our equal opportunity analysis concluded that White people had the most advantage and Hispanic had slight advantage, while being Asian gave you significant disadvantage and being Black gave you slight disadvantage.''}
This points to the tool successfully accomplishing user need \textbf{N4}.

\section{Lessons from an iterative design process}

Our foundational research revealed that WIT's target users typically rely on cumbersome model-specific code to conduct their analyses.
Our initial design focused on a single user journey in which users could--with no coding--edit a datapoint and re-run prediction. Over the course of 15 months, we added new interaction capabilities and refined the interface to better support other common tasks.

To that end, we streamlined WIT's functionality into three core workflows. In addition to the original workflow of testing hypothetical scenarios, we supported (a) general sense-making around data and (b) evaluation of performance and fairness. A key challenge in designing interfaces for these tasks was that many users were encountering certain components (e.g., partial dependence plots) and concepts (e.g., counterfactuals) for the first time. To address this issue, we provided copious in-application documentation. 

Feedback from users also led us to an interesting and counterintuitive design decision: to \textit{reduce} certain types of interactivity. When working with controls in the Fairness + Performance view, we initially made a direct connection between classification threshold controls and the data points visualization. A change in the threshold would lead to immediate rearrangement of items in the visualizations--a type of tight linking that is conventionally viewed as effective design. However, our users told us they found the constant changes (as well as an increased latency) confusing. To address this issue, we replaced the data points visualization (in this tab) with the confusion matrices and ROC curves. This simpler and more stable view proved much more satisfactory.

\section{Conclusion and Directions for Future Research}

We have presented the What-If Tool (WIT), which is designed to let ML practitioners explore and probe ML models. WIT enables users to analyze ML system performance on real data and hypothetical scenarios via a graphical user interface. The set of provided visualizations allows users to see a statistical overview of input data and model results, then zoom in to perform intersectional analysis. The tool lets users experiment with hypothetical conditions, with a variety of options that range from direct editing of data points to a novel type of automatic identification of counterfactual examples. It also provides capabilities for  assessing and optimizing metrics related to ML fairness, again without any coding.

Real-world usage of the tools indicates that these features are valuable not only to machine learning novices, but also to experienced ML engineers, and can help highlight issues with models that would be otherwise hard to see. In practical use we have seen WIT uncover nontrivial, long-lived bugs, and provide practitioners with new insights into their  models.

One natural future direction is to enhance the tool to make use of information about the internals of an ML model, when available. For instance, for models which represent differentiable functions, information about gradients can be informative about the saliency of different features \cite{sundararajan2017axiomatic} \cite{smilkov2017smoothgrad}. For feedforward neural networks, techniques such as TCAV \cite{kim2018tcav} can help lay users understand a model's output.

A particularly important future direction is to decrease the level of ML and data science expertise necessary to use this type of tool. For example, some users have suggested automating the process of finding outliers and subsets of data on which a model underperforms. Additionally, some users have requested to ability to add their own model performance metrics and fairness constraints into the tool through the UI. These and other directions have the potential to greatly broaden the set of stakeholders able to participate in ML model understanding and fairness efforts.

\acknowledgments{
The authors wish to thank the Google Research Big Picture and PAIR groups for their help throughout the entire process, along with all of the users from industry and academia that have used the tool and provided feedback.}

\bibliographystyle{abbrv}
\bibliography{template}

\begin{thebibliography}{10}

\bibitem{bpd}
Bpd field interrogation and observation dataset.
\newblock \url{https://data.boston.gov/dataset/boston-police-department-fio}.

\bibitem{facets}
Facets.
\newblock \url{https://pair-code.github.io/facets/}.

\bibitem{tableau}
Tableau.
\newblock \url{https://www.tableau.com/}.

\bibitem{tensorboard}
Tensorboard.
\newblock \url{https://www.tensorflow.org/tensorboard}.

\bibitem{tensorflow}
M.~Abadi, A.~Agarwal, P.~Barham, E.~Brevdo, Z.~Chen, C.~Citro, G.~S. Corrado,
  A.~Davis, J.~Dean, M.~Devin, S.~Ghemawat, I.~Goodfellow, A.~Harp, G.~Irving,
  M.~Isard, Y.~Jia, R.~Jozefowicz, L.~Kaiser, M.~Kudlur, J.~Levenberg,
  D.~Man\'{e}, R.~Monga, S.~Moore, D.~Murray, C.~Olah, M.~Schuster, J.~Shlens,
  B.~Steiner, I.~Sutskever, K.~Talwar, P.~Tucker, V.~Vanhoucke, V.~Vasudevan,
  F.~Vi\'{e}gas, O.~Vinyals, P.~Warden, M.~Wattenberg, M.~Wicke, Y.~Yu, and
  X.~Zheng.
\newblock {TensorFlow}: Large-scale machine learning on heterogeneous systems,
  2015.
\newblock Software available from tensorflow.org.

\bibitem{modeltracker}
S.~Amershi, M.~Chickering, S.~Drucker, B.~Lee, P.~Simard, and J.~Suh.
\newblock Modeltracker: Redesigning performance analysis tools for machine
  learning.
\newblock In {\em Proceedings of the Conference on Human Factors in Computing
  Systems (CHI 2015)}. ACM - Association for Computing Machinery, April 2015.

\bibitem{aif360-oct-2018}
R.~K.~E. Bellamy, K.~Dey, M.~Hind, S.~C. Hoffman, S.~Houde, K.~Kannan,
  P.~Lohia, J.~Martino, S.~Mehta, A.~Mojsilovic, S.~Nagar, K.~N. Ramamurthy,
  J.~Richards, D.~Saha, P.~Sattigeri, M.~Singh, K.~R. Varshney, and Y.~Zhang.
\newblock {AI Fairness} 360: An extensible toolkit for detecting,
  understanding, and mitigating unwanted algorithmic bias, Oct. 2018.

\bibitem{pmlr-v81-buolamwini18a}
J.~Buolamwini and T.~Gebru.
\newblock Gender shades: Intersectional accuracy disparities in commercial
  gender classification.
\newblock In {\em Proceedings of the 1st Conference on Fairness, Accountability
  and Transparency}, volume~81 of {\em Proceedings of Machine Learning
  Research}, pages 77--91, 2018.

\bibitem{crenshaw2018demarginalizing}
K.~Crenshaw.
\newblock Demarginalizing the intersection of race and sex: A black feminist
  critique of antidiscrimination doctrine, feminist theory, and antiracist
  politics [1989].
\newblock In {\em Feminist legal theory}, pages 57--80. Routledge, 2018.

\bibitem{DoshiKim2017Interpretability}
F.~Doshi-Velez and B.~Kim.
\newblock Towards a rigorous science of interpretable machine learning.
\newblock In {\em eprint arXiv:1702.08608}, 2017.

\bibitem{Dua2019}
D.~Dua and C.~Graff.
\newblock {UCI} machine learning repository, 2017.

\bibitem{hohman2018}
R.~P. D. H.~C. Fred~Hohman, Minsuk~Kahng.
\newblock Visual analytics in deep learning: An interrogative survey for the
  next frontiers.
\newblock In {\em IEEE Transactions on Visualization and Computer Graphics
  (TVCG)}, 2018.

\bibitem{hardt2016equality}
M.~Hardt, E.~Price, N.~Srebro, et~al.
\newblock Equality of opportunity in supervised learning.
\newblock In {\em Advances in neural information processing systems}, pages
  3315--3323, 2016.

\bibitem{gamut-a-design-probe-to-understand-howdata-scientists-understand-machine-learning-models}
F.~Hohman, A.~Head, R.~Caruana, R.~DeLine, and S.~Drucker.
\newblock Gamut: A design probe to understand how data scientists understand
  machine learning models.
\newblock ACM, May 2019.

\bibitem{compas}
S.~M. L.~K. J.~Angwin, J.~Larson.
\newblock Machine bias: There’s software used across the country to predict
  future criminals. and it’s biased against blacks.
\newblock {\em ProPublica}, May 2016.

\bibitem{kim2018tcav}
B.~Kim, M.~Wattenberg, J.~Gilmer, C.~Cai, J.~Wexler, F.~Viegas, and R.~Sayres.
\newblock Interpretability beyond feature attribution: Quantitative testing
  with concept activation vectors (tcav).
\newblock In {\em International Conference on Machine Learning}, pages
  2673--2682, 2018.

\bibitem{krause2017workflow}
J.~Krause, A.~Dasgupta, J.~Swartz, Y.~Aphinyanaphongs, and E.~Bertini.
\newblock A workflow for visual diagnostics of binary classifiers using
  instance-level explanations.
\newblock In {\em 2017 IEEE Conference on Visual Analytics Science and
  Technology (VAST)}, pages 162--172. IEEE, 2017.

\bibitem{prospector}
J.~Krause, A.~Perer, and K.~Ng.
\newblock Interacting with predictions: Visual inspection of black-box machine
  learning models.
\newblock In {\em Proceedings of the 2016 CHI Conference on Human Factors in
  Computing Systems}, CHI '16, pages 5686--5697, New York, NY, USA, 2016. ACM.

\bibitem{lim_2009}
B.~Y. Lim, A.~K. Dey, and D.~Avrahami.
\newblock Why and why not explanations improve the intelligibility of
  context-aware intelligent systems.
\newblock In {\em Proceedings of the SIGCHI Conference on Human Factors in
  Computing Systems}, CHI '09, pages 2119--2128, New York, NY, USA, 2009. ACM.

\bibitem{cnnvis}
M.~Liu, J.~Shi, Z.~Li, C.~Li, J.~Zhu, and S.~Liu.
\newblock Towards better analysis of deep convolutional neural networks.
\newblock {\em IEEE transactions on visualization and computer graphics},
  23(1):91--100, 2017.

\bibitem{DBLP:journals/corr/abs-1807-06228}
Y.~Ming, H.~Qu, and E.~Bertini.
\newblock Rulematrix: Visualizing and understanding classifiers with rules.
\newblock {\em CoRR}, abs/1807.06228, 2018.

\bibitem{tfx}
A.~N. Modi, C.~Y. Koo, C.~Y. Foo, C.~Mewald, D.~M. Baylor, E.~Breck, H.-T.
  Cheng, J.~Wilkiewicz, L.~Koc, L.~Lew, M.~A. Zinkevich, M.~Wicke, M.~Ispir,
  N.~Polyzotis, N.~Fiedel, S.~E. Haykal, S.~Whang, S.~Roy, S.~Ramesh, V.~Jain,
  X.~Zhang, and Z.~Haque.
\newblock Tfx: A tensorflow-based production-scale machine learning platform.
\newblock In {\em KDD 2017}, 2017.

\bibitem{NIST}
M.~Ngan.
\newblock {\em Face recognition vendor test (FRVT) performance of automated
  gender classification algorithms}.
\newblock 2015.

\bibitem{patel2008investigating}
K.~Patel, J.~Fogarty, J.~A. Landay, and B.~Harrison.
\newblock Investigating statistical machine learning as a tool for software
  development.
\newblock In {\em Proceedings of the SIGCHI Conference on Human Factors in
  Computing Systems}, pages 667--676. ACM, 2008.

\bibitem{pearl2014probabilistic}
J.~Pearl.
\newblock {\em Probabilistic reasoning in intelligent systems: networks of
  plausible inference}.
\newblock Elsevier, 2014.

\bibitem{auditai}
Pymetrics.
\newblock Audit ai.
\newblock \url{https://github.com/pymetrics/audit-ai}.

\bibitem{Ribeiro:2016:WIT:2939672.2939778}
M.~T. Ribeiro, S.~Singh, and C.~Guestrin.
\newblock "why should i trust you?": Explaining the predictions of any
  classifier.
\newblock In {\em Proceedings of the 22Nd ACM SIGKDD International Conference
  on Knowledge Discovery and Data Mining}, KDD '16, pages 1135--1144, New York,
  NY, USA, 2016. ACM.

\bibitem{smilkov2017smoothgrad}
D.~Smilkov, N.~Thorat, B.~Kim, F.~Vi\'{e}gas, and M.~Wattenberg.
\newblock Smoothgrad: removing noise by adding noise, 2017.

\bibitem{DBLP:journals/corr/StrobeltGHPR16}
H.~Strobelt, S.~Gehrmann, B.~Huber, H.~Pfister, and A.~M. Rush.
\newblock Visual analysis of hidden state dynamics in recurrent neural
  networks.
\newblock {\em CoRR}, abs/1606.07461, 2016.

\bibitem{sundararajan2017axiomatic}
M.~Sundararajan, A.~Taly, and Q.~Yan.
\newblock Axiomatic attribution for deep networks, 2017.

\bibitem{2009-ensemblematrix}
J.~Talbot, B.~Lee, A.~Kapoor, and D.~Tan.
\newblock Ensemblematrix: Interactive visualization to support machine learning
  with multiple classifiers.
\newblock In {\em ACM Human Factors in Computing Systems (CHI)}, 2009.

\bibitem{drawnet}
A.~Torralba.
\newblock Drawnet.
\newblock \url{http://brainmodels.csail.mit.edu/dnn/drawCNN/}.

\bibitem{wachter2017counterfactual}
S.~Wachter, B.~Mittelstadt, and C.~Russell.
\newblock Counterfactual explanations without opening the black box: automated
  decisions and the gdpr.
\newblock {\em Harvard Journal of Law \& Technology}, 31(2):2018, 2017.

\bibitem{Zhang_2019}
J.~Zhang, Y.~Wang, P.~Molino, L.~Li, and D.~S. Ebert.
\newblock Manifold: A model-agnostic framework for interpretation and diagnosis
  of machine learning models.
\newblock {\em IEEE Transactions on Visualization and Computer Graphics},
  25(1):364–373, Jan 2019.

\bibitem{zhaoiforest}
X.~Zhao, Y.~Wu, D.~Lee, and W.~Cui.
\newblock iforest: Interpreting random forests via visual analytics.
\newblock {\em IEEE Transactions on Visualization and Computer Graphics},
  PP:1--1, 09 2018.

\bibitem{pivot}
Y.~Zhao.
\newblock Pivot viewer based visualization of information analysis.
\newblock 01 2012.

\end{thebibliography}

\end{document}